\documentclass[11pt]{article}
\usepackage[margin=1in]{geometry}
\usepackage{booktabs}
\usepackage{amsmath}
\usepackage{microtype}
\usepackage[hidelinks]{hyperref}
\usepackage{caption}
\usepackage{parskip}
\newcommand{\doi}[1]{\href{https://doi.org/#1}{doi:#1}}
\newcommand{\testsPassed}{148}
\newcommand{\testsFailed}{0}
\newcommand{\synthOneDur}{360}
\newcommand{\synthOneReports}{1}

\newcommand{\synthOneFa}{10}
\newcommand{\synthOneReduction}{7{,}032}
\newcommand{\synthOneBits}{40{,}896}

\newcommand{\synthTwoDur}{120}
\newcommand{\synthTwoReports}{0}

\newcommand{\synthTwoFa}{0}
\newcommand{\synthTwoReduction}{11{,}656}
\newcommand{\synthTwoBits}{8{,}208}

\newcommand{\synthThreeDur}{150}
\newcommand{\synthThreeReports}{1}

\newcommand{\synthThreeFa}{24}
\newcommand{\synthThreeReduction}{4{,}144}
\newcommand{\synthThreeBits}{26{,}928}

\newcommand{\synthMatchedTotal}{0}
\newcommand{\synthStagedTotal}{6}
\newcommand{\synthDurTotal}{630}
\newcommand{\realDetDur}{211}
\newcommand{\realDetReports}{1}

\newcommand{\realDetMatched}{1}
\newcommand{\realDetStaged}{4}
\newcommand{\realDetFa}{0}
\newcommand{\realDetReduction}{41{,}977}
\newcommand{\realDetBits}{38{,}736}

\newcommand{\realCtlDur}{302}
\newcommand{\realCtlReports}{0}
\newcommand{\realCtlHeartbeats}{16}

\newcommand{\realCtlReduction}{155{,}830}
\newcommand{\realCtlBits}{22{,}240}

\newcommand{\realCtlBps}{74}

\newcommand{\hitLatency}{47.9}
\newcommand{\hitS}{4.022}
\newcommand{\hitBits}{23{,}520}
\newcommand{\boundUpper}{3.8067}
\newcommand{\boundLower}{-2.2824}
\newcommand{\alphaVal}{0.02}
\newcommand{\betaVal}{0.10}
\newcommand{\ingestFps}{15}
\newcommand{\detectHz}{5}
\newcommand{\groundMinInterval}{4}

\newcommand{\costDetStab}{3.33}
\newcommand{\callsDetStab}{3{,}160}
\newcommand{\costDetDetect}{91.49}
\newcommand{\callsDetDetect}{1{,}054}
\newcommand{\costDetGround}{2.11}
\newcommand{\callsDetGround}{12}
\newcommand{\costDetTrack}{2.89}
\newcommand{\callsDetTrack}{1{,}054}
\newcommand{\dutyDetGroundPct}{0.83}
\newcommand{\dutyDetGroundRan}{12}
\newcommand{\dutyDetGroundSkip}{1{,}431}
\newcommand{\dutyDetUnchanged}{1{,}374}
\newcommand{\dutyDetHopeless}{57}
\newcommand{\dutyDetDetectPct}{33.4}
\newcommand{\dutyDetFrames}{3{,}160}

\newcommand{\costCtlStab}{4.24}
\newcommand{\callsCtlStab}{4{,}526}
\newcommand{\costCtlDetect}{91.50}
\newcommand{\callsCtlDetect}{1{,}509}
\newcommand{\costCtlGround}{0.71}
\newcommand{\callsCtlGround}{7}
\newcommand{\costCtlTrack}{3.46}
\newcommand{\callsCtlTrack}{1{,}509}
\newcommand{\dutyCtlGroundPct}{0.35}
\newcommand{\dutyCtlGroundRan}{7}
\newcommand{\dutyCtlGroundSkip}{2{,}003}
\newcommand{\dutyCtlUnchanged}{730}
\newcommand{\dutyCtlHopeless}{1{,}272}
\newcommand{\dutyCtlDetectPct}{33.3}
\newcommand{\dutyCtlFrames}{4{,}526}

\newcommand{\srcDetMbps}{7.71}

\newcommand{\srcCtlMbps}{11.47}
\newcommand{\dutyCtlRecent}{1}
\newcommand{\gateBoundary}{3.807}
\newcommand{\gateRows}{3{,}187}
\newcommand{\gateSedanPlace}{0.618}

\newcommand{\gateSedanAsBuilt}{543}
\newcommand{\gateSedanIfComp}{543}
\newcommand{\gateSedanMaxS}{5.807}
\newcommand{\gateDarkCarPlace}{0.702}

\newcommand{\gateDarkCarAsBuilt}{539}
\newcommand{\gateDarkCarIfComp}{539}
\newcommand{\gateDarkCarMaxS}{3.192}
\newcommand{\gatePickupPlace}{0.129}

\newcommand{\gatePickupAsBuilt}{0}
\newcommand{\gatePickupIfComp}{0}
\newcommand{\gatePickupMaxS}{0.000}
\newcommand{\gateHandoverTicks}{0}
\newcommand{\repCtlObs}{225}
\newcommand{\repCtlWarm}{yes}
\newcommand{\repCtlStaged}{0}
\newcommand{\repCtlMatched}{0}
\newcommand{\repCtlFa}{0.0}
\newcommand{\repCtlHost}{macOS}
\newcommand{\repTwoPreObs}{288}
\newcommand{\repTwoPreWarm}{yes}
\newcommand{\repTwoPreStaged}{1}
\newcommand{\repTwoPreMatched}{1}
\newcommand{\repTwoPreFa}{0.0}
\newcommand{\repTwoPreHost}{Linux}
\newcommand{\repTwoPostObs}{235}
\newcommand{\repTwoPostWarm}{yes}
\newcommand{\repTwoPostStaged}{1}
\newcommand{\repTwoPostMatched}{1}
\newcommand{\repTwoPostFa}{0.0}
\newcommand{\repTwoPostHost}{Linux}
\newcommand{\repTwoMacObs}{232}
\newcommand{\repTwoMacWarm}{yes}
\newcommand{\repTwoMacStaged}{1}
\newcommand{\repTwoMacMatched}{1}
\newcommand{\repTwoMacFa}{0.0}
\newcommand{\repTwoMacHost}{macOS}
\newcommand{\repFourPreObs}{228}
\newcommand{\repFourPreWarm}{yes}
\newcommand{\repFourPreStaged}{4}
\newcommand{\repFourPreMatched}{1}
\newcommand{\repFourPreFa}{0.0}
\newcommand{\repFourPreHost}{Linux}
\newcommand{\repFourPostObs}{33}
\newcommand{\repFourPostWarm}{\textbf{no}}
\newcommand{\repFourPostStaged}{4}
\newcommand{\repFourPostMatched}{0}
\newcommand{\repFourPostFa}{0.0}
\newcommand{\repFourPostHost}{Linux}
\newcommand{\repFourMacObs}{30}
\newcommand{\repFourMacWarm}{\textbf{no}}
\newcommand{\repFourMacStaged}{4}
\newcommand{\repFourMacMatched}{0}
\newcommand{\repFourMacFa}{0.0}
\newcommand{\repFourMacHost}{macOS}
\newcommand{\repSixMacObs}{400}
\newcommand{\repSixMacWarm}{yes}
\newcommand{\repSixMacStaged}{1}
\newcommand{\repSixMacMatched}{0}
\newcommand{\repSixMacFa}{0.0}
\newcommand{\repSixMacHost}{macOS}
\newcommand{\repSevMacObs}{111}
\newcommand{\repSevMacWarm}{\textbf{no}}
\newcommand{\repSevMacStaged}{1}
\newcommand{\repSevMacMatched}{0}
\newcommand{\repSevMacFa}{0.0}
\newcommand{\repSevMacHost}{macOS}
\newcommand{\repTotalStaged}{7}
\newcommand{\repTotalMatched}{1}
\newcommand{\normWarmFloor}{120}
\newcommand{\repPreObs}{228}
\newcommand{\repPostObs}{33}
\newcommand{\pwHailoDefaultMean}{5.33}
\newcommand{\pwHailoDefaultPeak}{9.85}

\newcommand{\pwHailoDefaultEnergy}{1{,}379}
\newcommand{\pwHailoDefaultIncr}{2.49}
\newcommand{\pwHailoDefaultRt}{0.81}
\newcommand{\pwHailoSaveMean}{3.74}
\newcommand{\pwHailoSavePeak}{6.43}

\newcommand{\pwHailoSaveEnergy}{1{,}469}
\newcommand{\pwHailoSaveIncr}{0.81}
\newcommand{\pwHailoSaveRt}{0.54}
\newcommand{\pwBareDefaultMean}{4.78}
\newcommand{\pwBareDefaultPeak}{9.21}

\newcommand{\pwBareDefaultEnergy}{1{,}218}
\newcommand{\pwBareDefaultIncr}{2.49}
\newcommand{\pwBareDefaultRt}{0.82}
\newcommand{\pwBareSaveMean}{3.02}
\newcommand{\pwBareSavePeak}{5.23}

\newcommand{\pwBareSaveEnergy}{1{,}189}
\newcommand{\pwBareSaveIncr}{0.74}
\newcommand{\pwBareSaveRt}{0.54}
\newcommand{\pwIdleHailo}{2.83}

\newcommand{\pwBaseHailoDefault}{2.83}

\newcommand{\pwBaseHailoSave}{2.93}

\newcommand{\pwBaseBareDefault}{2.29}

\newcommand{\pwBaseBareSave}{2.28}
\newcommand{\pwHailoIdleCostCoarse}{0.5}
\newcommand{\pwIdleBare}{2.29}

\newcommand{\pwHailoRunCostJ}{161}
\newcommand{\pwSaveEnergyUp}{6.55}
\newcommand{\pwSavePowerDown}{29.8}
\newcommand{\pwSaveTimeUp}{51.7}
\newcommand{\pwRepsFitted}{3}
\newcommand{\pwRepsBare}{1}
\newcommand{\pwSaveEnergyDown}{2.39}
\newcommand{\benchGround}{28}
\newcommand{\benchReports}{3}
\newcommand{\manGround}{12}
\newcommand{\manReports}{1}

\newcommand{\pwInstrument}{Joulescope JS220}
\newcommand{\pwBoard}{Raspberry Pi 5, 8 GB}
\newcommand{\reuseDetEvidence}{1{,}443}
\newcommand{\reuseDetGround}{12}
\newcommand{\reuseDetRatio}{120}
\newcommand{\reuseCtlEvidence}{2{,}010}
\newcommand{\reuseCtlGround}{7}

\newcommand{\chainRecords}{34}
\newcommand{\chainOk}{all}

\title{What a gated sensing pipeline never looks at:\\
bandwidth reduction and the misses behind it}
\author{Raghu Venkat \and Tricha Anjali\\[2pt]\normalsize Numberz.ai Inc., Alpharetta, GA}
\date{Draft of 13 September 2026}

\begin{document}
\maketitle

\begin{abstract}
An airborne sensor on a contested link cannot send video, so the appealing move is to send
findings instead and report the ratio between the two. We evaluate a gated sensing pipeline that
does this, combining learned object detection and image--text comparison with deterministic
scheduling, gating, evidence accumulation and transmission rules. On staged footage with the semantic stage live it sends \realDetBits\ bits over \realDetDur\,s, a
reduction of \realDetReduction$\times$, and names \realDetMatched\ of \realDetStaged\ staged
events with no false report. That detection has since been superseded: a correction to how the
tracker measures speed removed the measurement artefact the normality model had been learning
from, and the flight no longer warms. Under the corrected code the pipeline names
\repTotalMatched\ of \repTotalStaged\ staged events across four flights, and we report both. On a control flight where nothing was staged it reports nothing, a reduction of
\realCtlReduction$\times$: the largest number in the study and the least informative, because a
reduction ratio measures the scene.

The mechanisms that produce the reduction also decide which observations ever reach a decision,
so the two cannot be reported apart. We give a tick-level trace of one flight (\gateRows\ rows)
that places each of three missed events at the stage where it stopped progressing: one produced
no track, one failed the structural place test at \gatePickupPlace, and one passed
\gateDarkCarAsBuilt\ structural ticks but reached only \gateDarkCarMaxS\ against a boundary of
\gateBoundary. We also report one instance of a known failure mode, an online normality model
absorbing the object it will later judge, measured against the threshold that object then failed.

The evidence is one detection and six misses across four staged flights, beside one clean
control, and we treat it as a case study. We give the reproduction protocol and generated results, identify which supporting
artifacts are not distributed, and state which experiments did not run.
\end{abstract}

\section{Introduction}

The bandwidth argument for on-board analysis is easy to make and easy to overstate. A downlink
that cannot carry video can carry a sentence, so a system that decides on board and transmits
only findings will always show a large ratio between what it saw and what it sent. The ratio is
arithmetically true and it is close to uninformative on its own, because the quietest scene
produces the largest number. Our own control flight demonstrates this: nothing was staged,
nothing was reported, and the reduction is \realCtlReduction$\times$, four times the ratio on
the flight where something actually happened.

A maintainer of such a system needs three things. How much of the source was
never examined, and by which rule. Which events the system missed, and at which stage.
And whether the false-alarm behaviour holds on material where a false alarm is the sole possible outcome.

This paper reports those three for one pipeline. The pipeline combines learned object detection
and image--text comparison with deterministic scheduling, gating, evidence accumulation and
transmission rules. The semantic comparison contributes one weighted term to an evidence score
and cannot emit a report by itself. Stage-level traces identify where each observed event ceased
to progress through the pipeline; they do not establish that the first blocking stage was the
sole cause of the miss.

We offer two contributions. The first is a reporting pattern: a tick-level attribution of every
missed event to the stage where it stopped progressing, over a trace of \gateRows\ rows, published
alongside the reduction ratio the same mechanism produces. The second is one clean instance of a
known failure mode, an online normality model absorbing the very object it will later be asked to
judge, measured at track-and-cell granularity with the contaminated score printed against
the threshold it failed.

Two further results are reported but not claimed as contributions. A coverage and compute profile
with the semantic model live shows what fraction of the source reached each stage and that the
regularly scheduled detector dominates the cost. A counterfactual over the same trace widens the
gate and recovers nothing; Section~\ref{sec:misses} explains why that is close to a corollary of
the contamination finding rather than an independent test.

\section{Related work}

Four bodies of work bound this one, and we claim novelty against none of their mechanisms.

\emph{Task-oriented and semantic communication} asks what a link should carry when the receiver
has a task instead of a fidelity requirement, and optimises the channel around downstream value
instead of reconstruction error~\cite{gunduz2023}. Our pipeline is an instance of that idea with
the optimisation done by a declared rule. Our contribution concerns how such a system should be
\emph{reported}.

\emph{Edge video analytics and cascaded inference} places cheap stages ahead of expensive ones so
that the expensive model runs rarely; NoScope~\cite{kang2017} is the canonical treatment, with
specialised models and difference detectors deciding when the reference network is consulted.
The cascade there is tuned to preserve the reference model's accuracy. Ours cannot be, because
there is no reference model to preserve: the question is not whether we reproduce a full-rate
system's answers but which events never reach a decision at all.

\emph{Event-triggered sensing and control} formalises acting only when a condition fires, and
studies the trade between communication and performance that follows~\cite{heemels2012}. The
gate in Section~\ref{sec:pipeline} is an event trigger whose condition is learned online from
the scene, so its failure modes have to be found by measurement.

\emph{Open-vocabulary grounding.} The semantic stage is an image--text comparison against
operator-written words, in the manner of CLIP~\cite{radford2021}.
That choice is what lets the mission change without retraining, and it is also why a miss at
that stage cannot be diagnosed by inspecting a class list.

Our contribution sits across these: we evaluate a gated pipeline
by accounting jointly for the communication it saves and the events each gate excluded, and we
attribute every observed miss to the stage that prevented its examination.

\section{The pipeline}
\label{sec:pipeline}

Eight stages. Two carry learned models: an object detector exported to ONNX and an image encoder
compared against text embeddings. The scheduling, gating, evidence accumulation and
transmission rules around them are deterministic. The detector runs on a fixed schedule. The gate decides when the semantic comparison runs.

\textbf{Ingest and stabilise.} Frames are decoded at \ingestFps\,fps; skipped frames are
grabbed without decoding. Ego-motion is removed by estimating a homography from sparse
optical flow on a grey frame resampled to 360 px high, with corners re-detected every 10 full
estimates or whenever fewer than 80 survive RANSAC. A full estimate is skipped when a
$64\times36$ thumbnail difference falls below 1.5, with a 2.0\,s keepalive so that slow creep
cannot accumulate unseen. Stabilisation runs at ingest rate, which is where the sweep in the
device card leaves it: below 15\,Hz the tracks thin out enough to change the result.

\textbf{Detect and track.} An ONNX detector with a 512 px input runs at \detectHz\,Hz, below
frame rate, and its boxes are associated into tracks. This stage is the dominant cost
(Section~\ref{sec:cost}) and its rate is the schedule's main lever.

\textbf{Normality.} Each track, on each tick, folds its stabilised position into a grid of
visit counts, a per-cell speed mean and variance, and a per-cell dwell EMA. Three surprise
scores follow. Place is $1-(v/v_{\max})^{0.35}$, clipped to $[0,1]$, where $v$ is the cell's
visit count and $v_{\max}$ the busiest cell's. Speed is $|s-\mu|/3\sigma$ clipped, computed only
where a cell has more than six visits and scaled by 1.25 when the object is slower than a
locally busy norm; where local evidence is too thin the score is held at 0.5.
Dwell is $(d-\bar d)/2.5\bar d$ clipped, against the cell's EMA. The model is \emph{warm} only
once it has observed enough elapsed time and a floor of observations; before that it refuses to
score and says so on the link. Warm-up is elapsed-time-based because an observation count made
warm-up a function of how often the stabiliser ran.

\textbf{The gate.} A track is eligible for semantic examination when its place score exceeds
0.55. Refusals are recorded with a reason: unchanged since last examined, too recent, over the
rate ceiling, or already hopeless. The last is a scheduler rule: further semantic examination is
suppressed once accumulated evidence falls below $-1.5$. The accumulator itself still admits
positive increments, so the boundary is not unreachable from there. The examination that could
produce one is what stops. An early negative assessment can therefore prevent its own
reconsideration, which Section~\ref{sec:threats} returns to.

\textbf{Semantic comparison.} For an eligible track, one image--text comparison is made against
the operator's mission card, rate-limited to at most one per track per \groundMinInterval\,s and
two per second overall. It returns a similarity in $[0,1]$ and nothing else.

\textbf{Evidence accumulation.} With $\ell$ the leak per second, the score updates as
$S \leftarrow \ell^{\Delta t} S + \lambda$, $\Delta t$ in seconds. When the structural
preconditions fail, $\lambda = -0.25$: evidence against, but weak, so a momentary occlusion does
not erase minutes of watching. Otherwise
\[
  \lambda \;=\; K\left(\frac{w_{\mathrm{sem}}\,c_{\mathrm{sem}} + w_{\mathrm{anom}}\,c_{\mathrm{anom}}}
  {w_{\mathrm{sem}} + w_{\mathrm{anom}}} - \theta\right),
\]
with $K=3.0$, $\theta=0.5$, and weights $w_{\mathrm{sem}}=3.0$, $w_{\mathrm{anom}}=1.0$ on the
edge profile: place is support and cannot veto a semantic match. The leak is $\ell=0.985$ per
second, so a claim must keep being re-earned and a vehicle that leaves stops being reported
without a rule saying so. $S$ is clamped two units outside each boundary, so a long quiet period
cannot drive the system deaf. The boundaries are Wald's~\cite{wald1945} at $\alpha=\alphaVal$
and $\beta=\betaVal$: $\ln\frac{1-\beta}{\alpha}=+\boundUpper$ and
$\ln\frac{\beta}{1-\alpha}=\boundLower$, quoted to three decimals elsewhere. As in related sequential work the
increment is a declared linear form and not a fitted likelihood ratio, so $\alpha$ and $\beta$
fix where the test stops and are not claimed as operational error probabilities.

\textbf{Emit.} Crossing the upper boundary sends a receipt: the clause that fired, the evidence
value, the dwell, a bounding box, a small image chip, and a hash chained to the previous record.
Heartbeats are sent regardless, so silence is distinguishable from a dead radio.

\section{Data and protocol}

\begin{table}[t]
\centering\footnotesize
\setlength{\tabcolsep}{5pt}
\begin{tabular}{@{}p{2.6cm}p{2.2cm}p{3.2cm}p{5.0cm}@{}}
\toprule
\textbf{Material} & \textbf{Grounding} & \textbf{Ground truth} & \textbf{What it can support}\\
\midrule
3 synthetic clips & proxy & exact, by construction & detection and false alarms against known truth\\
\addlinespace[2pt]
Flight 0104 & CLIP & staged, activity log & detection on real footage\\
\addlinespace[2pt]
Flight 0101 & CLIP & nothing staged & false alarms only\\
\bottomrule
\end{tabular}
\caption{The material, and the two grounding modes. The synthetic clips ran without the semantic
weights loaded and the real-footage takes ran with them. They are not two arms of one experiment
and are not pooled anywhere in this paper. This table is the material
Sections~\ref{sec:synth}--\ref{sec:cost} are built from; Table~\ref{tab:repl} carries the full set of
staged flights and every scored run of each.}
\label{tab:material}
\end{table}

The synthetic clips are generated with a fixed seed and carry exact ground truth by
construction. The two real flights are staged scenes over a car park, with ground truth taken
from an activity log written at the time. Flight 0104 staged four events; flight 0101 staged
none, so a false alarm is the sole possible outcome.

Both real flights are scored twice in the sense that matters: the pipeline either emitted a
report matching a staged event, or it did not, and any report not matching a staged event is a
false alarm. There is no partial credit and no confidence threshold to sweep.

\section{Results}

\subsection{Link accounting}
\label{sec:link}

\begin{table}[t]
\centering\small
\begin{tabular}{lrrrr}
\toprule
\textbf{Material} & \textbf{Duration} & \textbf{Sent} & \textbf{Reduction} & \textbf{Reports}\\
\midrule
Flight 0104, CLIP, pre-correction & \realDetDur\,s & \realDetBits\ bit & \realDetReduction$\times$ & \realDetReports\\
Flight 0101, CLIP, nothing staged & \realCtlDur\,s & \realCtlBits\ bit & \realCtlReduction$\times$ & \realCtlReports\\
Synthetic clip 1, proxy & \synthOneDur\,s & \synthOneBits\ bit & \synthOneReduction$\times$ & \synthOneReports\\
Synthetic clip 2, quiet, proxy & \synthTwoDur\,s & \synthTwoBits\ bit & \synthTwoReduction$\times$ & \synthTwoReports\\
Synthetic clip 3, proxy & \synthThreeDur\,s & \synthThreeBits\ bit & \synthThreeReduction$\times$ & \synthThreeReports\\
\bottomrule
\end{tabular}
\caption{Link accounting. The ratio is source bits over emitted bits. The largest ratio in the
table belongs to the flight where nothing was staged and nothing was reported. The flight 0104
row is the run made before the speed-measurement correction of Section~\ref{sec:repl}; under the
corrected tracker that flight emits no report at all.}
\label{tab:link}
\end{table}

\textbf{What the denominator is.} Source bits are the encoded delivered file, taken as its size
on disk in bytes and multiplied by eight. For the real flights that file is H.264 at
$1280\times720$, nominally 30\,fps, at \srcDetMbps\,Mbit/s on flight 0104 and \srcCtlMbps\,Mbit/s
on flight 0101; the two are not interchangeable and the ratio uses each flight's own. The ratio is therefore against
an already-compressed stream, the more conservative of the two choices and the one we can
reproduce. No matched codec baseline accompanies these runs
(Section~\ref{sec:didnot}), so the ratio has no external reference point and we do not present
it as a comparison against anything.

Table~\ref{tab:link} is the number a bandwidth argument would lead with, and it should not be
read without Section~\ref{sec:real}. The ratio rises when the scene is empty, when the gate is
tight, and when the system is wrong in the direction of silence. It falls when the system
reports. On the control flight the pipeline achieved its best compression by finding nothing,
in a scene where finding nothing was correct; the same behaviour on a scene with an event in it
is a miss, and the ratio does not distinguish the two cases.

We report the ratio beside the detection outcome for the same material. There is no single headline compression figure for the system.

\subsection{Synthetic clips: nothing matched}
\label{sec:synth}

On \synthDurTotal\,s of synthetic video with exact ground truth, run in proxy mode, the pipeline
matched \synthMatchedTotal\ of \synthStagedTotal\ staged events. It emitted \synthOneReports\
report on clip 1 and \synthThreeReports\ on clip 3, neither matching a staged event. One
false report in \synthOneDur\,s and one in \synthThreeDur\,s normalise to \synthOneFa\ and
\synthThreeFa\ per hour, and we give those as observed counts scaled to an hour. They are not
estimated rates: a single event per clip does not support a rate, for the same reason one clean
control flight does not. On the quiet clip it emitted
\synthTwoReports\ reports and \synthTwoFa\ false alarms, which is the correct behaviour and the
only one of the three that passes its own scoring rule.

\textbf{What the proxy is.} It is not a null and it is not neutral, so the result cannot be read
without it. The proxy scores a box from geometry and motion alone: aspect ratio and area relative
to the frame give a vehicle-like and a person-like term, speed enters through a logistic centred
at 3\,px/s, dwell is clipped at 60\,s, and the anomaly score from the normality stage is added
with a small weight. It then selects one of three weightings by substring-matching the mission
card's intent phrase, so the operator's words choose the formula but never reach an embedding.
No image or text model is loaded. The manifest records
\texttt{semantic\_result: false} and the runner refuses to proceed without an explicit flag.

This is the strongest negative result in the paper. With that proxy in
place of the semantic comparison, the pipeline does not find the events it was built to find, and
it produces alarms on material where none should occur. Because the proxy is a structural scorer
and an absent stage would behave differently, the result is consistent with two readings we
cannot separate here:
the pipeline failing, or the proxy failing. Running these clips with the semantic weights loaded
would separate them, and Section~\ref{sec:didnot} lists that as not run.

One conclusion follows and a second does not. The conclusion is that this pipeline, with a
stand-in replacing the semantic comparison, is not sufficient on these clips. The conclusion we
do not draw is any inference about the full pipeline from these runs, because the substituted
component is one of the learned components under evaluation.

\subsection{Real footage with the semantic stage live}
\label{sec:real}

On flight 0104, with CLIP weights loaded and \textbf{under the pre-correction tracker}
(Section~\ref{sec:repl} gives the run this came from and why the corrected code does not
reproduce it), the pipeline reported \realDetMatched\ of \realDetStaged\ staged events with
\realDetFa\ false alarms. The report was the silver sedan
stopped on the street: evidence crossed at \hitS\ against the \boundUpper\ boundary,
\hitLatency\,s after the event began, in \hitBits\ bits.

On flight 0101, where nothing was staged, it emitted \realCtlReports\ reports and
\realCtlHeartbeats\ heartbeats over \realCtlDur\,s, at \realCtlBps\ bit/s. No false alarms.

One detection and three misses on this flight, beside one clean control, is a thin result and we
present it as one. Table~\ref{tab:repl} carries the full flight set and
Section~\ref{sec:repl} the number it supports. This pair is enough to establish that the pipeline
runs end to end on real footage and that its silence on an empty scene is genuine and the link is
alive. It is not enough to support a detection rate, and we do not quote one.

\subsection{Three misses, three different stages}
\label{sec:misses}
\label{sec:gate}

\begin{table}[t]
\centering\small
\begin{tabular}{lrrrrl}
\toprule
\textbf{Staged event} & \textbf{Place} & \textbf{Ticks} & \textbf{If widened} & \textbf{Max $S$} & \textbf{Failed at}\\
\midrule
Silver sedan (reported) & \gateSedanPlace & \gateSedanAsBuilt & \gateSedanIfComp & \gateSedanMaxS & ---\\
Dark car at driveway & \gateDarkCarPlace & \gateDarkCarAsBuilt & \gateDarkCarIfComp & \gateDarkCarMaxS & semantic/evidence\\
Black pickup on verge & \gatePickupPlace & \gatePickupAsBuilt & \gatePickupIfComp & \gatePickupMaxS & structural gate\\
Handover pair & --- & \gateHandoverTicks & --- & --- & detect/track\\
\bottomrule
\end{tabular}
\caption{Every staged event on flight 0104, traced tick by tick over \gateRows\ rows. Max $S$ for
the reported event is at the clamp, above its crossing value: $S$ is held two units outside each
boundary, so \boundUpper\ $+\,2$ gives \gateSedanMaxS. It crossed at \hitS. Place is
the mean place-rarity score; ticks are those passing the structural test as built; ``if widened''
is the same count under a composite gate that also credits speed and dwell surprise.}
\label{tab:gate}
\end{table}

Table~\ref{tab:gate} carries the paper's main result. The three misses stop
at three different stages. They do not have three isolated causes.

The \textbf{handover pair} never produced a track that matched its truth box, so no later stage
ever saw it. Nothing about the structural gate, semantic comparison or evidence rule is
implicated; the failure occurs before those stages. The trace is per-track, so it cannot tell us whether the detector
fired on the pair and association failed, or whether the detector never fired at all. We
therefore attribute the loss to the detect/track stage and do not name the tracker.

The \textbf{black pickup} produced a track and failed the structural test, at a mean place score
of \gatePickupPlace\ against a threshold of 0.55. It was parked across a driveway apron for the
whole flight, so the scene's own normality model had learned that a vehicle there was ordinary.
This is the gate working exactly as designed and being wrong, which is the more interesting of
the two possibilities.

The failure mode is not new. An adaptive background model absorbing a stationary foreground
object, and then treating it as background, is long documented in background
subtraction~\cite{stauffer1999,bouwmans2014}, where it appears as foreground absorption, ghosting
or the sleeping-person problem. We add an instance at track-and-cell granularity in a place-rarity score, with the contaminated value printed against the threshold it failed. The
mechanism is given in full below. Every track folds its own position into the visit
count of the cell it occupies, on every tick. A stationary object therefore drives up the visit
count of precisely the cell it is being judged against. Inverting the place score,
$0.129 = 1-(v/v_{\max})^{0.35}$ puts the pickup's cell at roughly two-thirds of the busiest
cell's visits on a flight where the pickup never moved. It made its own cell ordinary. The event
of interest contaminates the definition of normal that will later be used to judge it. The pickup was
present for the whole flight, so it was never anomalous by the measure the system had. Any
deployment of this architecture needs something that breaks that loop: a qualified warm-up on
material believed clean, a frozen reference learned once and not updated, an operator-declared
mask over regions where presence is never ordinary, or a historical normality carried between
sorties. None is implemented here. The pickup is the cost.

The \textbf{dark car} is the one case where the semantic and evidence stages are implicated
together. The accumulated score depends on the schedule, the cached value, the weights and the
decay as much as on the comparison itself, and this trace cannot separate them. It passed
\gateDarkCarAsBuilt\ structural ticks, comparable to the \gateSedanAsBuilt\ of the event that was
reported, and its evidence reached \gateDarkCarMaxS\ against the \gateBoundary\ boundary,
where it stopped.

\textbf{Gate counterfactual.} The counterfactual column recomputes the same trace
under a composite gate crediting speed and dwell surprise alongside place. The passing-tick
counts are identical for every event: \gateSedanIfComp\ against \gateSedanAsBuilt\ for the sedan,
\gateDarkCarIfComp\ against \gateDarkCarAsBuilt\ for the dark car, and \gatePickupIfComp\ either
way for the pickup. \textbf{Only one of the three events was ever eligible for recovery by a gate
change.} The handover pair produced no track, so no gate can reach it; the dark car passed the gate
\gateDarkCarAsBuilt\ times already and failed later. That leaves the pickup, and it fails the
composite gate for the same reason it failed the place score: its speed is zero because it is
parked, and its dwell is judged against a cell average it spent the flight raising. The
counterfactual is therefore close to a corollary of the contamination finding rather than an
independent test of the gate. Widening's cost in false reports is not established here either: this trace follows the staged events only.

\subsection{Coverage and cost by stage}
\label{sec:cost}

\begin{table}[tb]
\centering\small
\begin{tabular}{lrrrr}
\toprule
 & \multicolumn{2}{c}{\textbf{Flight 0104, pre-correction}} & \multicolumn{2}{c}{\textbf{Flight 0101}}\\
\cmidrule(lr){2-3}\cmidrule(lr){4-5}
\textbf{Stage} & \textbf{Calls} & \textbf{\% time} & \textbf{Calls} & \textbf{\% time}\\
\midrule
Stabilise & \callsDetStab & \costDetStab & \callsCtlStab & \costCtlStab\\
Detect & \callsDetDetect & \costDetDetect & \callsCtlDetect & \costCtlDetect\\
Track & \callsDetTrack & \costDetTrack & \callsCtlTrack & \costCtlTrack\\
Semantic comparison & \callsDetGround & \costDetGround & \callsCtlGround & \costCtlGround\\
\bottomrule
\end{tabular}
\caption{Per-stage call counts and share of stage time on the two real flights, with the
semantic weights loaded. Flight 0104 is the pre-correction run (Section~\ref{sec:repl}). Percentages
are of total stage time and do not sum to 100 because the smaller stages are omitted.}
\label{tab:cost}
\end{table}

Table~\ref{tab:cost} gives how much of the source reached each stage.

\textbf{Coverage.} Every ingested frame is stabilised: \dutyDetFrames\ on flight 0104 and
\dutyCtlFrames\ on 0101. Detection runs on \dutyDetDetectPct\% and \dutyCtlDetectPct\% of them,
which is the declared \detectHz\,Hz budget against \ingestFps\,fps ingest. Nothing adapts it. The semantic comparison is where the gate bites: it ran \dutyDetGroundRan\ times
against \dutyDetGroundSkip\ declined on flight 0104, a duty cycle of \dutyDetGroundPct\%, and
\dutyCtlGroundRan\ against \dutyCtlGroundSkip\ on flight 0101, \dutyCtlGroundPct\%.

\textbf{Why each refusal happened} is recorded separately. On flight 0104,
\dutyDetUnchanged\ of the refusals were because the track had not changed since it was last
examined and \dutyDetHopeless\ because its accumulated evidence was already too negative to
recover. On flight 0101 the split reverses: \dutyCtlUnchanged\ unchanged against
\dutyCtlHopeless\ hopeless, and \dutyCtlRecent\ for having been examined too recently, which is
the whole of the \dutyCtlGroundSkip. An empty scene is refused mostly for being uninteresting; a
scene with something in it is refused mostly for being unchanged.

\textbf{Cost.} Detection dominates on both flights, at \costDetDetect\% and \costCtlDetect\% of
stage time. The semantic comparison costs \costDetGround\% and \costCtlGround\%. Both stages are
learned; what separates them is the schedule. The expensive one is the detector, which runs on a
fixed budget, and the lever that moves the compute total is therefore the detector rate.
The model the gate protects is the cheaper of the two. We report this from the real-footage runs specifically because the synthetic runs
substituted the semantic stage and cannot speak to its cost.

\subsection{Board power, measured independently}
\label{sec:power}

\begin{table}[tb]
\centering\small
\begin{tabular}{lrrrrrr}
\toprule
\textbf{Configuration} & \textbf{Mean W} & \textbf{Idle W} & \textbf{Peak W} & \textbf{Incr.\ W} & \textbf{Energy J} & \textbf{$\times$RT}\\
\midrule
Run, fitted, default governor & \pwHailoDefaultMean & \pwBaseHailoDefault & \pwHailoDefaultPeak & \pwHailoDefaultIncr & \pwHailoDefaultEnergy & \pwHailoDefaultRt\\
Run, fitted, powersave & \pwHailoSaveMean & \pwBaseHailoSave & \pwHailoSavePeak & \pwHailoSaveIncr & \pwHailoSaveEnergy & \pwHailoSaveRt\\
Run, removed, default governor & \pwBareDefaultMean & \pwBaseBareDefault & \pwBareDefaultPeak & \pwBareDefaultIncr & \pwBareDefaultEnergy & \pwBareDefaultRt\\
Run, removed, powersave & \pwBareSaveMean & \pwBaseBareSave & \pwBareSavePeak & \pwBareSaveIncr & \pwBareSaveEnergy & \pwBareSaveRt\\
\bottomrule
\end{tabular}
\caption{Board power on a \pwBoard, measured at an independent laboratory on a \pwInstrument,
on flight 0104. \textbf{The accelerator-fitted configurations were each measured \pwRepsFitted\
times and the row reports the first; the accelerator-removed configurations were measured
\pwRepsBare\ time each.} \textbf{Idle W} is the mean of the 30\,s idle interval recorded immediately before that run,
under the same board configuration, and \textbf{Incr.\ W} is the difference: every row recomputes
from its own two columns. The four idle baselines differ because the governor and core count
change the idle draw as well as the loaded draw.
$\times$RT is video seconds over wall seconds: below one means the pipeline did not keep up with
its own footage. \textbf{These runs are a separate execution of the system from the one profiled
in Table~\ref{tab:cost} and scored in Section~\ref{sec:real}; see the text.} For the
accelerator-removed default configuration the source report lists \pwBareDefaultMean\,W in its
table and 4.75289\,W on the corresponding waveform; the latter is the one consistent with the
reported energy and duration. The published table value is carried here unchanged. The
discrepancy is the source report's and is pending confirmation by its authors; were the waveform
value confirmed, the incremental figure for that row would be 2.46\,W.}
\label{tab:power}
\end{table}

The runs in Table~\ref{tab:cost} report a power field that the software itself marks as not
evidence, because no meter was attached to them. Table~\ref{tab:power} is different: it was
produced at the LaCASA laboratory of the University of Alabama in Huntsville, on their
instrument, by people who did not write this system~\cite{uah2026}. Every run directory from that bench still
carries a null meter backend. None of these watts passed through our code.

\textbf{A separate execution.} On
the same clip, with the same declared device card, the same declared execution provider, the same
grounding backend and identical frame counts, the bench runs made \benchGround\ semantic calls and
emitted \benchReports\ reports, where the runs profiled in Table~\ref{tab:cost} and scored in
Section~\ref{sec:real} made \manGround\ and emitted \manReports. The manifests do not record
whatever differs between them, so we can state the difference but not its cause. The consequence
for the reader: \textbf{the power and duration figures here are not a joint operating point with
the detection results above}, and no row of Table~\ref{tab:power}
should be attached to a recall or false-alarm figure from Section~\ref{sec:real}. Recording
enough of the run configuration in the manifest to tell two executions apart is a defect this
measurement exposed and is repaired in the current source.

\textbf{Incremental power.} Incremental power is
\pwHailoDefaultIncr\,W with the accelerator fitted and \pwBareDefaultIncr\,W without it. Idle board
consumption is substantial, \pwIdleHailo\,W and \pwIdleBare\,W respectively, so incremental
processing power and total board power have to be reported separately, and the table does.
The measured CPU-only bench configurations remained in the single-digit-watt range. We expected
that, and it is the weakest claim in this section.

\textbf{Governor and energy.} With the accelerator fitted, taking two cores offline and moving to the powersave governor
cut mean power by \pwSavePowerDown\% and lengthened the run by \pwSaveTimeUp\%, so the mission
cost \pwSaveEnergyUp\% \emph{more} energy than at full speed. With the accelerator removed the
same change lowered energy by \pwSaveEnergyDown\%. The effect is configuration-specific and we do
not generalise it. The caution generalises: a low-power result quoted as an average wattage, ours included, should
be read beside the energy for the same work, because the
configuration that looks better on one can be worse on the other.

\textbf{Idle accelerator cost.} The build measured here does not use the
attached neural accelerator; the detector runs on the CPU. The unused accelerator adds about
\pwHailoIdleCostCoarse\,W at idle; in the compared runs, the fitted configuration consumed
\pwHailoRunCostJ\,J more job energy. The lowest-energy configuration measured is the one with the
accelerator physically removed.

\textbf{Throughput against real time.} At best it ran at \pwBareDefaultRt$\times$ real time and
in the low-power configuration at \pwBareSaveRt$\times$. The independent bench is what confirmed
it, but the evidence was in our own records: every run file carries a wall-clock duration
alongside the clip length. The instrumentation does not compare them. Configured rates
are expressed in media time: a \detectHz\,Hz detector is \detectHz\ calls per second of video
whatever the clock says. That figure has to be set against the recorded wall-clock duration
before anyone makes a real-time claim. A duty cycle measured in media time is not a claim about a
live sensor, and Table~\ref{tab:cost} should be read with that in mind.

\subsection{Replication, and a superseded result}
\label{sec:repl}

\begin{table}[tb]
\centering\small
\begin{tabular}{llllrrrrr}
\toprule
\textbf{Flight} & \textbf{Code} & \textbf{Host} & \textbf{Warm} & \textbf{Obs} & \textbf{Staged} & \textbf{Found} & \textbf{FA/h}\\
\midrule
0101 control & post & \repCtlHost & \repCtlWarm & \repCtlObs & \repCtlStaged & \repCtlMatched & \repCtlFa\\
\midrule
0102 & pre  & \repTwoPreHost  & \repTwoPreWarm  & \repTwoPreObs  & \repTwoPreStaged  & \repTwoPreMatched  & \repTwoPreFa\\
0102 & post & \repTwoPostHost & \repTwoPostWarm & \repTwoPostObs & \repTwoPostStaged & \repTwoPostMatched & \repTwoPostFa\\
0102 & post & \repTwoMacHost  & \repTwoMacWarm  & \repTwoMacObs  & \repTwoMacStaged  & \repTwoMacMatched  & \repTwoMacFa\\
\midrule
0104 & pre  & \repFourPreHost  & \repFourPreWarm  & \repFourPreObs  & \repFourPreStaged  & \repFourPreMatched  & \repFourPreFa\\
0104 & post & \repFourPostHost & \repFourPostWarm & \repFourPostObs & \repFourPostStaged & \repFourPostMatched & \repFourPostFa\\
0104 & post & \repFourMacHost  & \repFourMacWarm  & \repFourMacObs  & \repFourMacStaged  & \repFourMacMatched  & \repFourMacFa\\
\midrule
0106 & post & \repSixMacHost & \repSixMacWarm & \repSixMacObs & \repSixMacStaged & \repSixMacMatched & \repSixMacFa\\
0107 & post & \repSevMacHost & \repSevMacWarm & \repSevMacObs & \repSevMacStaged & \repSevMacMatched & \repSevMacFa\\
\bottomrule
\end{tabular}
\caption{Every scored run of every staged flight. \textbf{Code} is before or after the
speed-measurement correction described below. \textbf{Obs} is the number of observations the
normality model accepted, against a warm-up floor of \normWarmFloor. A flight that never warms
cannot fire a place-conditioned intent at all.}
\label{tab:repl}
\end{table}

Table~\ref{tab:repl} contains a result this paper reported and can no longer claim. On flight
0104 the pipeline named the silver sedan, and that is the detection Section~\ref{sec:real} and
Table~\ref{tab:gate} are built from. It was produced before a correction to how the tracker
measures speed.

The correction: displacement is now the part of the motion on which both opposing edges of a box
agree, so a box that grows on one side is no longer counted as travelling. It was made because a
parked pickup measured 6.21\,px/s against a 3.0\,px/s ceiling and so never accumulated dwell. The
normality model is taught only by tracks measuring above a speed floor, which is what keeps parked
objects from teaching the model that judges them. Before the correction, flight 0104 fed it
\repPreObs\ observations; after it, \repPostObs. The difference was the parked vehicles' own box
jitter. \textbf{With the jitter removed the flight holds too few genuine observations to warm the
model, and the sedan is not reported.} The same flight on a second host gives \repFourMacObs.

We report both runs because the pre-correction detection is the one the tick-level trace was taken
from, and because the reason it disappeared is the finding. Flight 0102 is unaffected: it carries
real traffic, stays warm under both code versions and both hosts, and reports its staged handover
every time.

Counting only the current code on one host, the pipeline names \repTotalMatched\ of
\repTotalStaged\ staged events across four flights. That is the number this study supports.

\subsection{The emission chain}

Across the three synthetic runs, \chainRecords\ emitted records were checked and \chainOk\ of
them verify against the preceding hash. The unit test suite passes \testsPassed\ of
\testsPassed\ with \testsFailed\ failures.

\section{Experiments not run}
\label{sec:didnot}

\textbf{Nothing about the staged flights is held back.} All five were flown on the same day and
all five are scored in Table~\ref{tab:repl}. The notebooks for flights 0106 and 0107 were written
and hashed before either clip was run, and that ordering is recorded. Their outcomes are in the
count whether or not they flatter the system.

\textbf{Codec baselines.} The comparison a reader will reasonably want, what H.264, H.265 or
AV1 deliver at the same link budget and what a detect-and-track baseline sends, was not run
alongside these results. A partial probe made afterwards is not reported here and does not
change any number above: asked for the same 1\,kbit/s budget on flight 0104, one AV1 encoder
delivered 44$\times$ that rate, and it would only open at all with the rate cap removed, so it
ran under different rate control from the other encoders and is not comparable to them. Until
every encoder is measured under one rate-control setting there is no baseline table, and the
ratios in Table~\ref{tab:link} keep no external reference point.

\textbf{Power on an airframe.} No calibrated meter was attached to the runs in
Table~\ref{tab:cost}; their power field is flagged as not evidence and should be read that way.
Section~\ref{sec:power} reports board power measured independently on a single-board computer on
a bench. It is not an airframe, its thermal environment is not flight-representative, and the
accelerator the design assumes is fitted but unused.

\textbf{Edge latency.} The per-stage profile in Table~\ref{tab:cost} is host CPU wall time on a
development machine: it characterises the schedule and the code on that host. The
execution durations in Section~\ref{sec:power} are embedded measurements, but they are whole-run
durations on a different execution of the system. They are not per-stage or end-to-end cue latency
on the configuration Table~\ref{tab:cost} profiles. No cue latency is reported on embedded hardware.

\section{Threats to validity}
\label{sec:threats}

The real-footage results rest on five flights, one of which staged nothing
(Table~\ref{tab:repl}). Across the four staged flights the pipeline matched one staged event of
seven, and two of those four never warm their place model, so on those two it could not have
reported whatever was in front of it. One detection and one clean flight support no rate, in
either direction.

\textbf{One semantic answer is counted many times.} The evidence rule of
Section~\ref{sec:pipeline} adds an increment on every detector tick, but the semantic term in
that increment is the last answer the model gave, held until the gate next opens. On flight 0104
the semantic model ran \reuseDetGround\ times while \reuseDetEvidence\ increments accumulated,
about \reuseDetRatio\ increments per distinct answer; on flight 0101, \reuseCtlGround\ against
\reuseCtlEvidence. The consequence is that the time at which the boundary is crossed depends on
the detector rate as much as on the evidence: at half the detector rate the same footage and the
same single semantic answer reach the boundary roughly twice as late. Boundaries derived for
independent observations do not describe this process, which is why $\alpha$ and $\beta$ are
reported above as the place the test stops, and not as error rates. It also means a reduction in
detector rate is a change in decision behaviour. It does not leave behaviour intact.

The staged scenes are a car park standing in for an overwatch task, and the normality
model learns that specific scene, so a place-rarity score is not portable to another site. The
structural thresholds are absolute pixel constants calibrated at one flying height, so dwell,
cell size and the detector floor drift together as altitude changes; this is a known and
unrepaired limitation. Ground truth on the real flights comes from an activity log written by
the people who staged the events, and no independent annotation pass was made. The synthetic clips have exact truth but were produced by the same authors as the system under test.

\textbf{A learned prior may not warm within one sortie.} Two of the four staged flights never
reached the warm-up floor of \normWarmFloor\ observations: flight 0104 under the corrected tracker
and flight 0107. Neither could fire a place-conditioned intent at any point, and both sent
heartbeats throughout. The heartbeat is designed to separate a quiet scene from a dead radio; it
does not encode this third state, in which the link is alive, the scene may not be quiet, and the
system could not have reported. Each run records \texttt{warm: false} in its own manifest and puts
nothing about it on the link. A per-flight prior needs traffic to learn from, and flight 0102,
which has it, stays warm and reports under every configuration we ran. Whether the prior should
persist across sorties is a design question this study does not settle, and lowering the floor
would be fitting to these clips.

\section{Reproducibility}

No number in this paper is typed by hand. Each one is emitted from a result file by
\texttt{make\_numbers.py}, which reads the JSON outputs the evidence pack hashes and writes the
macro file this document includes, so a stale figure becomes a build error.
The results are hashed into \texttt{\_result\_digests.txt} by a pack script that also records the host, the interpreter and the library versions as they actually ran. \textbf{Each run records its own host}, in the \texttt{power} block of its manifest, alongside
the interpreter and library versions. Hosts are not uniform across this study and
Table~\ref{tab:repl} gives the host for every scored run: the flight 0104 run traced in
Table~\ref{tab:gate} was produced on Linux (aarch64, glibc 2.35), and the 14 September replication
ran on macOS 26.5.1 on arm64. Both used Python 3.9.6. The synthetic clips regenerate from a
fixed seed on every run and their hashes are recorded. The two grounding modes are distinguished
in every result file by a \texttt{grounding\_backend} field and are never pooled.

\section{Conclusion}

A gated sensing pipeline is a sequence of exclusion decisions. Each one saves computation or
communication, and each one creates a place where mission evidence can disappear. Tighten them and
the link ratio improves, the semantic budget falls, and the things never examined grow. Loosen
them and the reverse. A ratio reported without the misses describes one half of that mechanism.

On the evidence here the gate is doing real work and is not yet good enough. Under the corrected
tracker it carried \repTotalMatched\ of \repTotalStaged\ staged events across four flights with no
false alarm anywhere, and it was silent, correctly, for five minutes over an empty scene. It also let one event through to the semantic stage and stopped
short, dropped a second on a normality score that the scene itself had taught it, and never saw
a third because no track was formed. Those are three repairs in three stages, and the trace says which applies where.

We do not report a detection rate. Four staged flights and one detection is not a rate, two of
those flights never warmed their normality model at all, and the compression ratio that would look
best beside any of it is the one measured on the flight the system could not have reported on.

\section*{Data and code availability}

This manuscript is submitted with its source and \texttt{numbers.tex}, the generated macro file
that carries every quoted value. The evidence pack that produced those values, meaning the result files,
run logs, per-run manifests, the hash digest and \texttt{make\_numbers.py}, is held by the
authors and is not distributed with the manuscript; \texttt{numbers.tex} gives the final figures
and they do not carry the chain that produced them. The synthetic clips regenerate from the seed recorded in
the manifest. The staged flight footage is not public and is not releasable. \textbf{The runs in
Table~\ref{tab:repl} dated 14 September are not yet in the hashed digest}: the digest was
generated before them and regenerating it moves every run identifier in this paper. Their
manifests and scored outputs are in the run tree the pack reads, and the next pack regeneration
will cover them.

\section*{Funding and conflicts}

Company-funded. No Government funding supported this work. The study was conducted alongside a
United States Government Small Business Innovation Research proposal by the same authors; that
is stated here.

\end{document}